\documentclass[letterpaper]{article} 
\usepackage{aaai25}
\usepackage{times}  
\usepackage{helvet}  
\usepackage{courier}  
\usepackage[hyphens]{url}  
\usepackage{graphicx} 
\usepackage{natbib}  
\usepackage{caption} 
\usepackage{algorithm}
\usepackage{algorithmic}
\usepackage{amssymb}
\usepackage{amsmath,bm}

\usepackage{booktabs}
\usepackage{makecell}
\usepackage{colortbl}
\usepackage{newfloat}
\usepackage{listings}
\DeclareCaptionStyle{ruled}{labelfont=normalfont,labelsep=colon,strut=off} 
\floatstyle{ruled}
\newfloat{listing}{tb}{lst}{}
\floatname{listing}{Listing}
\newcommand{\textlink}[2]{%
  \begingroup
  \urlstyle{same}
  #2%
  \endgroup
}

\title{Large Images are Gaussians: High-Quality Large Image \\ Representation with Levels of 2D Gaussian Splatting
}
\author{
    Lingting Zhu\textsuperscript{\rm 1}, Guying Lin\textsuperscript{\rm 2}, Jinnan Chen\textsuperscript{\rm 3}, Xinjie Zhang\textsuperscript{\rm 4}, \\
    Zhenchao Jin\textsuperscript{\rm 1}, Zhao Wang\textsuperscript{\rm 5}, Lequan Yu\textsuperscript{\rm 1 {\footnote{Corresponding author.}}} 
}

\affiliations{
   \textsuperscript{\rm 1} The University of Hong Kong, Hong Kong SAR, China\\
   \textsuperscript{\rm 2} Carnegie Mellon University, PA, USA\\
   \textsuperscript{\rm 3} National University of Singapore, Singapore\\
   \textsuperscript{\rm 4} The Hong Kong University of Science and Technology, Hong Kong SAR, China\\
   \textsuperscript{\rm 5} The Chinese University of Hong Kong, Hong Kong SAR, China\\
   ltzhu99@connect.hku.hk, lqyu@hku.hk
}

\begin{document}

\maketitle

\begin{abstract}
While Implicit Neural Representations (INRs) have demonstrated significant success in image representation, they are often hindered by large training memory and slow decoding speed. Recently, Gaussian Splatting (GS) has emerged as a promising solution in 3D reconstruction due to its high-quality novel view synthesis and rapid rendering capabilities, positioning it as a valuable tool for a broad spectrum of applications. In particular, a GS-based representation, 2DGS, has shown potential for image fitting. In our work, we present \textbf{L}arge \textbf{I}mages are \textbf{G}aussians (\textbf{LIG}), which delves deeper into the application of 2DGS for image representations, addressing the challenge of fitting large images with 2DGS in the situation of numerous Gaussian points, through two distinct modifications: 1) we adopt a variant of representation and optimization strategy, facilitating the fitting of a large number of Gaussian points; 2) we propose a Level-of-Gaussian approach for reconstructing both coarse low-frequency initialization and fine high-frequency details. Consequently, we successfully represent large images as Gaussian points and achieve high-quality large image representation, demonstrating its efficacy across various types of large images. 
Code is available at {\textlink{https://github.com/HKU-MedAI/LIG}{https://github.com/HKU-MedAI/LIG}}.
\end{abstract}

\section{Introduction}

Existing researches have challenged the prevailing assumption that images are best represented as uniform pixel grids given the continuous nature of real visual world. Traditional signal processing methods, such as Discrete Cosine Transform (DCT)~\cite{khayam2003discrete}, which transfers the spatial signal into the frequency domain, have been effectively applied to lossy image compression techniques, \textit{e.g.}, JPEG~\cite{rabbani2002overview}. With the rapid advancement of neural networks, which have demonstrated remarkable efficiency in function approximation~\cite{lecun2015deep}, researchers are increasingly turning to neural representations for a wide range of fitting-based applications. This shift is central to the field of representation learning~\cite{bengio2013representation}. In the realm of image representation, a notable example is the Local Implicit Image Function~\cite{chen2021learning}, which employs Implicit Neural Representations (INRs)~\cite{chen2019learning, sitzmann2020implicit, park2019deepsdf} to map continuous coordinates to their corresponding signals at any resolution. INR-based methods typically use a compact neural network to produce an implicit continuous mapping, preserving intricate image details and opening up new possibilities for applications such as image compression and super-resolution~\cite{dupont2021coin,chen2021learning, ma2022recovering, strumpler2022implicit}.

\begin{figure}[t]
\centering
\includegraphics[width=1.00\columnwidth]{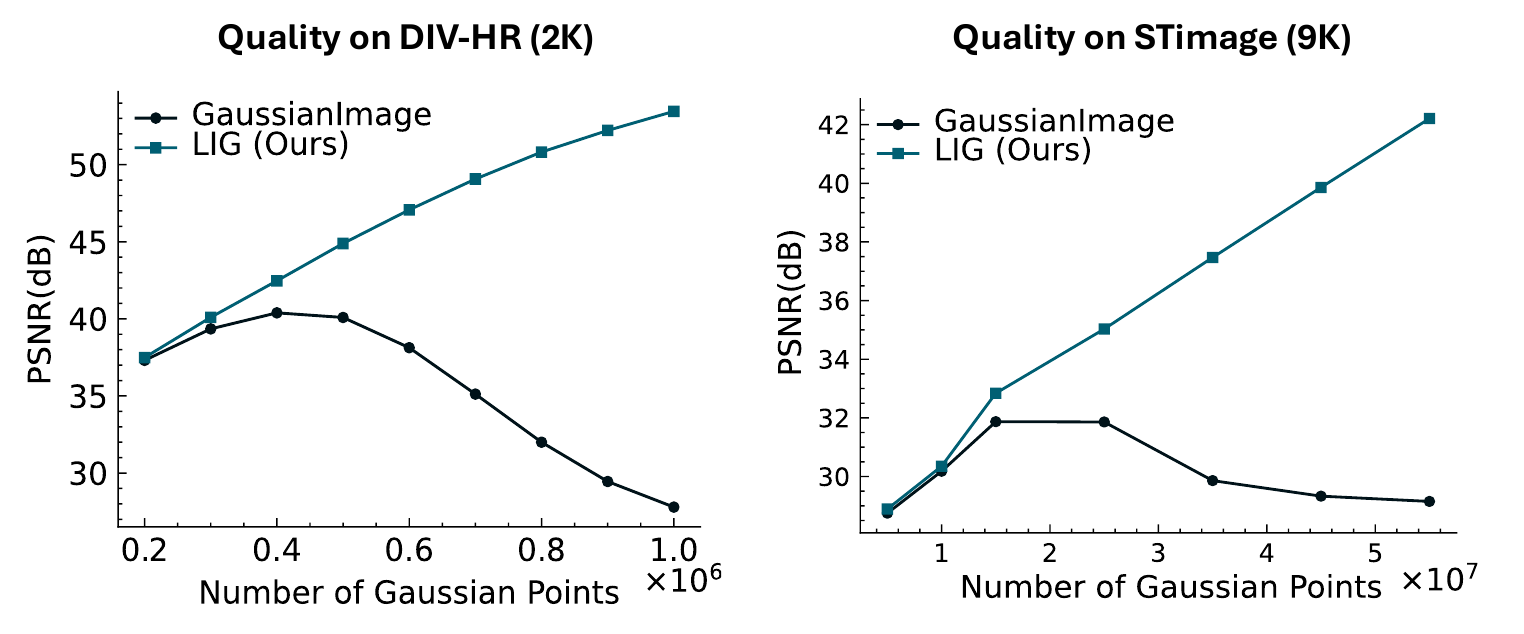}
\caption{\textbf{Comparison of LIG and GaussianImage on large image fitting quality.} GaussianImage performs badly when optimizing a large number of Gaussian points on images of high resolutions, whereas ours consistently delivers quality improvements as the number of Gaussian points increases. The phenomenon is observed in multiple datasets.}
\label{varying_points}
\end{figure}

\begin{figure*}[t]
\centering
\includegraphics[width=1.00\textwidth]{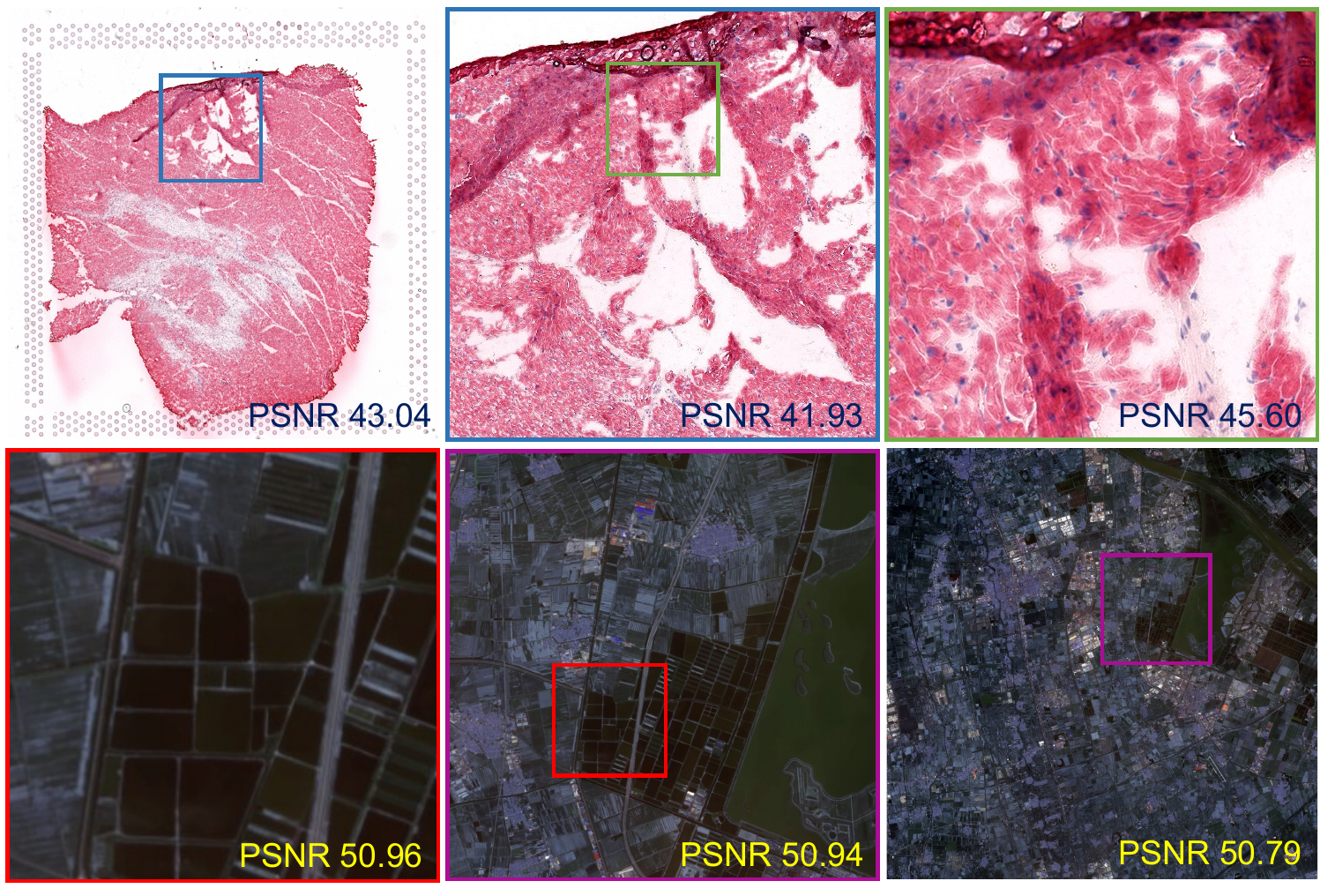}
\caption{\textbf{LIG is capable of representing large images with high quality.} We show cases including a histopathology image and a satellite image, showing multi-resolution patches with PSNR values displayed at the bottom-right corner of each image.} 
\label{teaser}
\end{figure*}

Despite their potential, most INR-based methods suffer from substantial training memory and slow decoding speed, largely due to their reliance on grid-based features~\cite{sitzmann2020implicit, saragadam2023wire, ramasinghe2022beyond, liu2024finer}. Specifically, the coordinates are of grid-like structures and mapped to neural features for Multi-Layer Perception (MLP) processing. This approach presents two significant drawbacks: 1) the grid-based structures, which grow quadratically, result in dramatically large equivalent batch sizes, bringing substantial or even prohibitive training memory requirements; 2) the decoding process, which necessitates parallel processing of neural networks, despite ongoing advancements in this area, can be slow for large batches. These limitations become particularly critical when dealing with larger target signals, \textit{e.g.,} large images, which is the primary focus of our work. 

3D Gaussian Splatting~\cite{kerbl20233d} (3DGS), designed for 3D scene reconstruction, has emerged as a novel representation known for its high-quality, real-time rendering capabilities. This is largely due to its use of explicit 3D Gaussians and differentiable tile-based rasterization~\cite{lassner2021pulsar}. In an effort to address the aforementioned challenges and explore the potential of Gaussian Splatting (GS) for image representation, GaussianImage~\cite{zhang2024gaussianimage} introduces a 2D Gaussian Splatting representation for images, advocating an image-space tailored rasterization method for efficient training and rendering. Similarly, Image-GS~\cite{zhang2024image} adaptively allocates and progressively optimizes 2D Gaussians for efficient representation learning. These pioneering works have demonstrated the effectiveness of GS for image fitting, achieving comparable quality and higher efficiency than INR-based methods for small images, while maintaining a satisfactory signal-to-noise ratio.

However, existing GS-based fitting methods have yet to demonstrate their potential for higher fidelity and their adaptability to large images. In our work, we delve deeper into the application of 2DGS representation for large images, which naturally require a larger number of Gaussian points compared to smaller images. We introduce \textbf{L}arge \textbf{I}mages are \textbf{G}aussians (\textbf{LIG}). As shown in Fig.~\ref{varying_points}, LIG is capable of fitting large images with an increasing number of Gaussian points, a task where GaussianImage may fall short. We examine the optimization difficulties encountered by GaussianImage when dealing with a large number of Gaussian points and, in response, we make two distinct modifications to overcome these challenges. Firstly, we optimize the Gaussian parameters using a slightly different 2DGS representation, aided by re-implemented CUDA kernels.
Specifically, we directly optimize the covariance matrix without decompositions which are managed in 3DGS and GaussianImage~\cite{kerbl20233d,zhang2024gaussianimage} and maintain semi-definiteness via post-processing. Secondly, we harness the concept of Level of Detail (LOD) from computer graphics, proposing a Level-of-Gaussian approach for hierarchically fitting large images. This shares similarities with research adopting LOD in Neural Field and Gaussian Splatting, including MINER, BungeeNeRF, Octree-GS, and Hierarchical 3D Gaussian~\cite{saragadam2022miner, ren2024octree, xiangli2022bungeenerf, kerbl2024hierarchical}. With the Level-of-Gaussian mechanism, we can allocate a small ratio of points for initializing coarse low-frequency structures, leaving high-frequency details for second-stage fitting with the majority of Gaussian points. Unlike MINER, which is related to the sensitivity of grid features and fixed resolution for input coordinates, we splatter all Gaussian points simultaneously for different Gaussian number configurations, and select the number of levels as 2 for images of any resolution. 

Our designs ease the training of numerous Gaussian points and enable GS-based representation for large images. As demonstrated in Fig.~\ref{teaser}, LIG can perform high-quality fitting for large images, where we select medical images and remote sensing images for illustration, highlighting the potential for applications in telemedicine and satellite communication~\cite{mittermaier2023digital, de2015satellite}. In summary, our main contributions are highlighted as follows:
\begin{itemize}
    \item We are the first to delve into applying GS-based representation for large images, aiming at reconstructing large images with Gaussian points.
    \item We make two designs on 2DGS for images, namely, 2DGS representation, and the utilization of a two-stage Level-of-Gaussian approach, which mitigate the training obstacles of a large number of Gaussians.
    \item We compare our method with baselines on various types of large images, including general visual high-resolution images, high-resolution histopathology images, and satellite images.
\end{itemize}
\section{Related Works}

\noindent\textbf{Implicit Neural Representations.}
Since early works focused on representing the signed distance field for 3D shapes~\cite{chen2019learning, park2019deepsdf, xu2019disn, michalkiewicz2019implicit}, Implicit Neural Representations (INRs) have been applied to a variety of applications involving different types of representations, including 3D scenes~\cite{mildenhall2021nerf, barron2021mip}, images~\cite{saragadam2022miner, dupont2021coin, strumpler2022implicit, chen2021learning}, and videos~\cite{chen2021nerv, chen2023hnerv, li2022nerv}. A notable example is NeRF~\cite{mildenhall2021nerf}, which uses a Multilayer Perceptron (MLP) to represent geometry and view-dependent appearance, sparking a surge of interest in 3D vision and graphics. Most INRs use grid-based input processing for spatial signals and focus more on activation functions like sinusoids~\cite{sitzmann2020implicit}, Gabor wavelets~\cite{saragadam2023wire}, and variable-periodic activation functions~\cite{liu2024finer} to capture high-quality details for fitting. However, these methods suffer from inefficient training and inference, making them unsuitable for high-resolution images. To handle large-scale signals, multi-resolution signal representations that rely on efficient feature querying or hierarchical processing have been widely adopted~\cite{muller2022instant, saragadam2022miner, martel2021acorn, chen2023neurbf}. Among them,  while hierarchical architectures, such as MINER~\cite{saragadam2022miner}, can facilitate large image fitting at the cost of sequential training, they do not address the fundamental issues associated with the grid-based nature of INRs, leading to prolonged training times and slow decoding speeds. Recent attempts to improve INRs have included the use of multi-resolution hash encoding or radial basis functions~\cite{muller2022instant, chen2023neurbf}, which have achieved high accuracy with fewer parameters. These advancements have significantly improved INRs, making them the current leading approaches.

\noindent\textbf{3D Gaussian Splatting.} 
3D Gaussian Splatting (3DGS)~\cite{kerbl20233d} becomes a trend in the computer graphics and computer vision communities due to its ability to compactly represent complex 3D scenes while enabling high-speed rendering. The scene is modeled as 3D Gaussian primitives, each of which is defined by position, scale, rotation, opacity, and appearance attributes. The parameters of the Gaussians are optimized to align observations via differentiable rendering.
3DGS shows remarkable potential across a wide range of downstream tasks. These include novel view synthesis and scene reconstruction~\cite{kerbl20233d,kerbl2024hierarchical,yu2024mip,huang20242d,zhu2024deformable,zhao2024hfgs,liu2024endogaussian,li2024endosparse}, 3D reconstruction and generation~\cite{tang2024lgm, xu2024grm, szymanowicz2024splatter, tang2023dreamgaussian, liu2024humangaussian, chen2024generalizable}, and applications in robotics~\cite{yugay2023gaussian, lu2024manigaussian,matsuki2024gaussian}.
Recent advancements in 3DGS further incorporate Level-of-Detail (LOD) techniques to enhance rendering efficiency and enable adaptive scene representation~\cite{ren2024octree,kerbl2024hierarchical,yan2024multi}, particularly crucial for large-scale scene reconstruction which requires visual quality with real-time rendering.

\noindent\textbf{Gaussian Splatting Based Image Representation.} 
Recently, GaussianImage~\cite{zhang2024gaussianimage} is the first to adapt 3DGS for image representations. Specifically, it adapted Gaussian points to image spaces with fewer characterizing parameters and employed alpha blending to merge color and opacity attributes. After per-sample image fitting, the attributes can be compressed using quantization-aware fine-tuning. As a result, GaussianImage can represent images with 2DGS and compress them while maintaining high quality. Image-GS~\cite{zhang2024image} also utilizes the eight parameters for 2D Gaussian points and fits a target image by adaptively allocating and progressively optimizing a set of 2D Gaussians. GaussianSR~\cite{hu2024gaussiansr}, assigns a learnable Gaussian kernel to each pixel for super-resolution.
Another related work, Splatter Image~\cite{szymanowicz2024splatter}, embeds Gaussian attributes at the image level, achieving ultra-efficient 3D reconstruction. In the field of image fitting, despite their groundbreaking success, it remains unclear whether GS can be used to fit large images at a quality that can compete with INRs. Given the nature of large fitting targets, it is necessary to allow for more Gaussian points to be optimized simultaneously. We demonstrate that GaussianImage falls short in this regard due to difficulties in optimizing their representations and the lack of multi-resolution mechanisms for capturing high-frequency information.
\section{Methodology}

In this section, we present our Large Images are Gaussians (LIG) framework. We begin with the basics of 3DGS and its adaptation to 2D spaces. Subsequently, we delve into two primary components of our methodology: 1) the variant of 2DGS representation; 2) the Level-of-Gaussian mechanism; which provide an efficient and high-quality solution for optimizing large images containing numerous Gaussian points.

\subsection{Preliminaries}
\noindent \textbf{3D Gaussian Splatting (3DGS).} 
As proposed by \cite{kerbl20233d}, 3D Gaussian splatting employs a set of 3D Gaussians to represent 3D scenes. Each Gaussian is characterized by its mean
\( \bm x \in \mathbb{R}^{3}\), scale \(\bm s \in \mathbb{R}^{3}\), rotation $\bm r \in \mathbb{R}^{3}$, opacity $\alpha \in \mathbb{R}$, and color $\bm c \in \mathbb{R}^{c}$. Spherical harmonics can be used to further define view-dependent effects. The rendering process involves projecting these 3D Gaussians onto the image plane, resulting in 2D elliptical splats and performing $\alpha$-blending for each pixel in a front-to-back depth order. Compared to neural rendering techniques like Neural Radiance Fields (NeRF)~\cite{mildenhall2021nerf}, 3DGS provides faster rendering and efficient training capabilities.

\noindent \textbf{2D Gaussian Splatting (2DGS).}
In~\cite{zhang2024gaussianimage}, the Gaussians are adapted to 2D spaces and are defined by deduced parameters. We only need to formulate the 2D Gaussian points locating on the fixed plane corresponding to the image. Specifically, each 2D Gaussian can be described by its position $\bm \mu \in \mathbb{R}^{2}$, 2D covariance matrix $\bm \Sigma \in \mathbb{R}^{2\times 2}$, color coefficients $\bm c \in \mathbb{R}^3$, and opacity $o \in \mathbb{R}$. To ensure the positive semi-definite, $\bm \Sigma$ can be decomposed with Cholesky factorization or into a rotation matrix $\bm R \in \mathbb{R}^{2\times 2}$ and scaling matrix $\bm S \in \mathbb{R}^{2\times 2}$ following~\cite{kerbl20233d}, which altogether requires 3 parameter in 2D cases. The $\alpha$-blending, which calculates the color of pixel $i$ via
\begin{equation}
  \begin{aligned}
    \boldsymbol{C}_i = \sum_{n \in \mathcal{N}} \boldsymbol{c}_n \cdot \alpha_n \cdot T_n, \quad T_n = \prod_{m=1}^{n-1} (1 - \alpha_m),
  \end{aligned}
\end{equation}
where $T_n$ represents the accumulated transparency. And $\alpha_n$ is computed with 2D covariance $\boldsymbol{\Sigma}$ and opacity $o_n$:
\begin{equation}
  \begin{aligned}
    \alpha_n = o_n \cdot \exp(-\sigma_n), \quad \sigma_n = \frac{1}{2} \boldsymbol{d}_n^T \boldsymbol{\Sigma}_n^{-1} \boldsymbol{d}_n,
  \end{aligned}
\end{equation} 
where $\boldsymbol{d} \in \mathbb{R}^2$ is the displacement between the pixel center and the projected 2D Gaussian center. Moreover, with the accumulated summation mechanism~\cite{zhang2024gaussianimage}, the opacity can be integrated into the representation of color, resulting in arbitrary-value $\bm c_n' \in \mathbb{R}^3$ to reduce the number of parameters. In total, we have 8 parameters for each 2D Gaussian, with 2 parameters for position, 3 parameters for covariance, 3 parameters for weighted colors. And the color of pixel $i$ is computed as $\boldsymbol{C}_i = \sum_{n \in \mathcal{N}} \boldsymbol{c}_n' \cdot \exp(-\sigma_n).$
Both \cite{zhang2024gaussianimage} and \cite{zhang2024image} employ the 8 parameters formulation. We keep the accumulated summation mechanism and use 8 parameters but only change the representation of the covariance matrix.

\subsection{2D Gaussians Formulation}
We adopt a variant of representation and optimization strategy on 2D Gaussians. Our representation on each point differs only in 2D covariance matrix $\bm \Sigma \in \mathbb{R}^{2\times 2}$. If we decompose this into rotation matrix $\bm R$ and scaling matrix $\bm S$, we have a total of 3 variables. After extensive experiments on exploring the capabilities of 2DGS, we find that optimizing the decomposed parameters can be challenging when the Gaussian points are numerous. Therefore, we opt for optimizing the covariance matrix directly, which also requires 3 parameters considering the symmetry of the covariance matrix. The forward and backward kernel functions in CUDA are accordingly implemented for optimization.

As stated in~\cite{kerbl20233d}, covariance matrices have physical meaning when they are positive semi-definite. Despite that we cannot ensure its positive semi-definite via directly optimizing the upper triangular of the matrix, we can have two important assertions in 2DGS cases. Firstly, the Gaussian points do not necessarily need to retain their physical meaning in image fitting where the representations are considered as the semi-implicit fitters. And the covariance is only used to produce $\sigma_n$ as the weight of the color with an exponential activation. Secondly, we can filter out Gaussian points with covariance matrices that do not obey positive semi-definite in 2D cases. In detail, when computing the color for a certain pixels, removing all $\sigma_n < 0$ can eliminate points with invalid covariance matrices. Consequently, the color changes to 
\begin{equation}
  \begin{aligned}
    \boldsymbol{C}_i = \sum_{n \in \mathcal{N}, \sigma_n > 0} \boldsymbol{c}_n' \cdot \exp(-\sigma_n).
  \end{aligned}
\end{equation}
It is guaranteed that for any positive semi-definite matrix with a non-zero determinant, the inverse matrix will also be positive semi-definite, which implies
\begin{equation}
  \begin{aligned}
\boldsymbol{\Sigma}_n \text{ is positive semi-definite, } \boldsymbol{\Sigma}_n^{-1} \text{ exists} \\ \rightarrow \boldsymbol{\Sigma}_n^{-1} \text{ is positive semi-definite} \rightarrow \sigma_n > 0, \forall \boldsymbol{d}_n
  \end{aligned}
\end{equation}
As the contrapositive also holds true, for each pixel, Gaussian points that we filter out are not positive semi-definite.

Yet with the above operation,  we cannot ensure that all covariance matrices producing positive $\sigma$ have physical meanings, since the $\boldsymbol{d}_n$ is drawn from a finite set. Given the nature of the 2D cases, it is straightforward to determine whether a 2D symmetric matrix is positive semi-definite. Matrices without physical meaning can be strictly filtered out if this is a desired feature. We have included the relevant proofs and analysis in Supplementary Material for reference.

\begin{figure}[h]
\centering
\includegraphics[width=1.00\columnwidth]{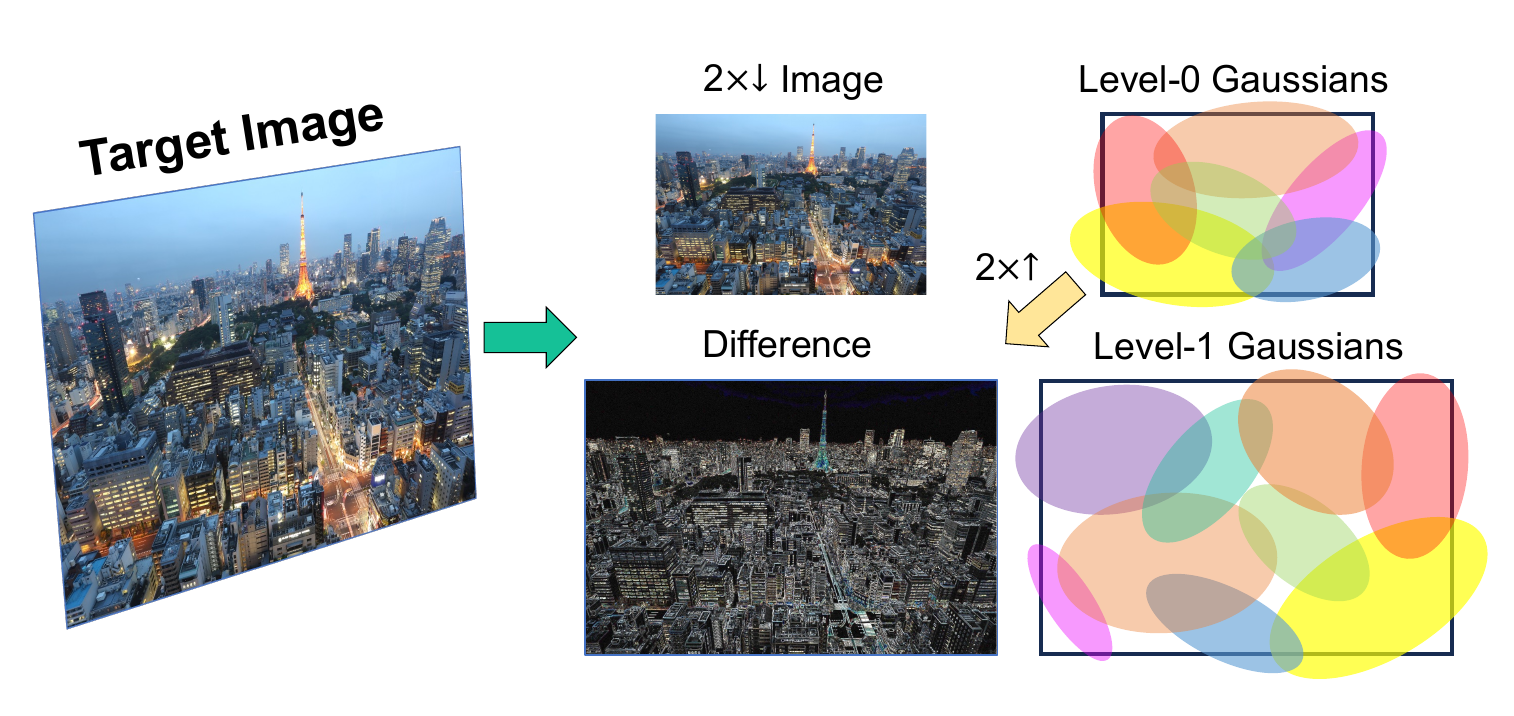}
\caption{\textbf{Illustration of our proposed Level-of-Gaussian approach, aiming at fitting large images with two levels of Gaussian points.} In the first stage, we allocate parts of Gaussian points to form $L_0$ Gaussians for learning the low-frequency initialization from the down-sampled image. In the second stage, $L_1$ Gaussians learn the high-frequency details on the difference between the up-sampled estimation and the target. We present the abstract values of the difference and enhance the image for visualization.}
\label{framework}
\end{figure}

\subsection{Levels of 2D Gaussians}
Fig.~\ref{framework} provides an illustration of the Level-of-Gaussian mechanism. The concept of Level of Detail, which refers to the complexity of a 3D model representation, has been widely used in computer graphics. Existing works, which share core principles with our design, employ multi-scale designs for high-resolution signals. For instance, BungeeNeRF~\cite{xiangli2022bungeenerf} develops a progressive NeRF-based model to fit large scenes, while MINER~\cite{saragadam2022miner} opts to use multiple MLPs to fit images at different levels. Our design addresses similar concerns, which can be summarized in two main points: 1) the models for different levels and 2) the target signals at different levels. 

Our objective is to fit large images using levels of Gaussians. Unlike MINER, which employs grid-based MLPs that are sensitive to input coordinates for a fixed resolution, our levels of Gaussians can efficiently splatter all points and remain robust for target resolutions. In contrast to NeRF-based methods that primarily use multi-resolution targets to facilitate the training of MLPs, our design on levels is more focused on forcing the Gaussians to fit the retargeted one, which can be low-frequency structure and high-frequency details. This approach also allows us to convert the large images at the final level into normalized targets, \textit{i.e.,} the difference image. Given the additional training and inference costs of sequential training, we set the level number as 2, where each level learns either low-frequency or high-frequency information. A detailed analysis on the selection of the level number can be found in Supplementary Material.

Given a target image $I$, we optimize the Levels of 2D Gaussian Splatting $\{L_0(\mathcal{N}_0, T_0), L_1(\mathcal{N}_1, T_1)\}$, each of which has the corresponding 2D Gaussians $\mathcal{N}_i$ and the fitting target $T_i$.
First, we consider fitting image targets at multiple resolutions as earlier works and the targets naturally require different number of Gaussian points, which aligns with the network sizes in the neural field. We allocate a portion of the Gaussian points to learn the low-frequency initialization of the target images from the down-sampled image. Since the training target of $L_0$ is of lower resolution and can be simpler than that of $L_1$, we assign fewer points on $L_0$. Consequently, given a total Gaussian number of the Levels of Gaussians as $|\mathcal{N}|$, we have 
\begin{align}
|\mathcal{N}| = |\mathcal{N}_0|+|\mathcal{N}_1| = (1+r)|\mathcal{N}_1|
\end{align} 
where $r$ is the allocation ratio, set as 0.125 in our experiments. After the first-stage tuning, the $L_0$ produces low-frequency initialization that is used to produce the fitting target of $L_1$. Since the rendering only produces images of values between 0 and 1, we normalize the values of difference image via a min-max scaler. The min and max values should be saved in the model for inference, taking only two values for better performance.  The target $T_0$ and $T_1$ can be expressed as:
\begin{align}
T_0 &= {\rm{Down}}(I),\\
T_1 &= {\rm{Norm}}(I-{\rm{UP}}({{\rm{Render}}(L_0)})).
\end{align} 

During training, we set the same number of iterations for the two levels. The fitting is supervised with Mean Squared Error (MSE) loss for each stage. Experiments demonstrate that the two levels of fitting significantly improve performance on large images and only mildly increase the expenses on time for training and testing, each of which involves two rounds of splatting. Additionally, the two levels of Gaussians also reduce the training memory. This is because the first level is frozen when the second one is training, leading to a reduction in maximum number of points with gradients given a fixed total number of Gaussians.

\section{Experiments}

\subsection{Experimental Setup}
\textbf{Dataset.} 
We assess our method across three diverse datasets, encompassing medical, remote sensing, and general visual tasks. These datasets vary in resolution where we utilize 15 9K histopathology images of human heart sourced from STimage~\cite{chen2024stimage1k4m}, 4 4K satellite images from the Full-resolution Gaofen-2 (FGF2)~\cite{wang2024cross}, and 2K DIV-HR dataset, comprising 100 images, with low-resolution counterparts evaluated in~\cite{zhang2024gaussianimage}.

\noindent \textbf{Evaluation Metrics.} We use three metrics to evaluate our method against GS-based state-of-the-art and INR-based counterparts. We employ PSNR for image quality, which quantify the distortion between reconstructed images and original images. As one merit of LIG is the fewer training memory for large images compared to GaussinImage and INRs, we provide training memory for reference. Benefiting from GS representation, our method achieves high rendering speed and FPS is used for benchmark.

\noindent \textbf{Implementation Details.} Since we present a new 2DGS representation for images, CUDA kernels are incorporated and we build the packages upon gsplat~\cite{ye2023mathematical}. The training steps for $L_0$ and $L_1$ are set to the same, 30,000 steps in our implementation. The learning rate is 0.018 and Adam optimizer is used~\cite{kingma2014adam}.

\subsection{Main Results}

\begin{table*}[ht]
  \centering
\resizebox{\textwidth}{!}{
  \begin{tabular}{lcccccccccc}
    \toprule
     & & \multicolumn{3}{c}{STimage (9K)} &  \multicolumn{3}{c}{FGF2 (4K)} & \multicolumn{3}{c}{DIV-HR (2K)} \\
    \cmidrule(r){3-5} \cmidrule(r){6-8} \cmidrule(r){9-11}
    Method & Reference & PSNR $\uparrow$ & Tr. Mem. (GB) $\downarrow$ & FPS $\uparrow$ & PSNR $\uparrow$ & Tr. Mem. (GB) $\downarrow$ & FPS $\uparrow$ & PSNR $\uparrow$ & Tr. Mem. (GB) $\downarrow$ & FPS $\uparrow$  \\
    \midrule
     \textbf{INR-based} & & & & & & & & & & \\
    \midrule
    SIREN & NeurIPS'20 & - & - & - & - & - &  - & 28.61 & 28.05 & 38.55 \\
    Gauss & ECCV'22 & - & - & - & - & - & - & 25.39 & 38.56 & 25.12 \\
    WIRE & CVPR'23 & - & - & - & - & - & - & 24.42 & 52.40 & 8.92 \\
    FINER & CVPR'24 & 17.74 & 71.27 & 12.81 & 21.91 & 74.82 & 12.35 & 34.42 & 80.01 & 9.84 \\
    \midrule
    \textbf{GS-based} & & & & & & & & & & \\
    \midrule
    GaussianImage + (3.5e7, 1e7, 5e5) & ECCV'24  & 29.86 & 20.47 & 19.86 & 27.50 & 5.39 & \textbf{78.19} & 40.09 & 1.03 & \textbf{745.01} \\
    GaussianImage + (4.5e7, 1.2e7, 7e5) & ECCV'24  & 29.33 & 23.56 & \textbf{18.95} & 27.53 & 5.64 & \textbf{69.24} & 35.12 & 1.09 & \textbf{635.30} \\
    GaussianImage + (5.5e7, 1.4e7, 9e5) & ECCV'24  & 29.28 & 25.89 & \textbf{16.51} & 27.48 & 6.04 & \textbf{67.20} & 29.45 & 1.17 & \textbf{525.73} \\
\rowcolor[rgb]{ .863,  .922,  1}  \textbf{LIG (Ours)} + (3.5e7, 1e7, 5e5) & - & \textbf{37.47} & \textbf{16.67} & \textbf{20.19} & \textbf{51.81} & \textbf{4.21 }& 74.38 & \textbf{44.89} & \textbf{1.01} & 541.84 \\
\rowcolor[rgb]{ .863,  .922,  1}   \textbf{LIG (Ours)} + (4.5e7, 1.2e7, 7e5) & - & \textbf{39.82} & \textbf{17.75} & 17.89 & \textbf{53.90} & \textbf{4.26} & 63.62 & \textbf{49.07} & \textbf{1.02} & 491.37 \\
\rowcolor[rgb]{ .863,  .922,  1}   \textbf{LIG (Ours)} + (5.5e7, 1.4e7, 9e5) & - & \textbf{42.19} & \textbf{20.26} & 15.72 & \textbf{56.05} & \textbf{4.39} & 58.09 & \textbf{52.22} & \textbf{1.05} & 441.76 \\
    \bottomrule
  \end{tabular}
}
  \caption{\textbf{Quantitative results on three datasets.} We report the PSNR, Training Memory, and FPS for all methods. For GS-based methods, we present the number of Gaussian points for each dataset, denoted in the form of tuples. Please note that for INR-based methods, the fitting may be infeasible for large image datasets due to the large memory. ``3.5e7'' denotes $3\times 10^7$.}
  \label{quantitative}
\end{table*}

\begin{figure*}[t]
\centering
\includegraphics[width=1.00\textwidth]{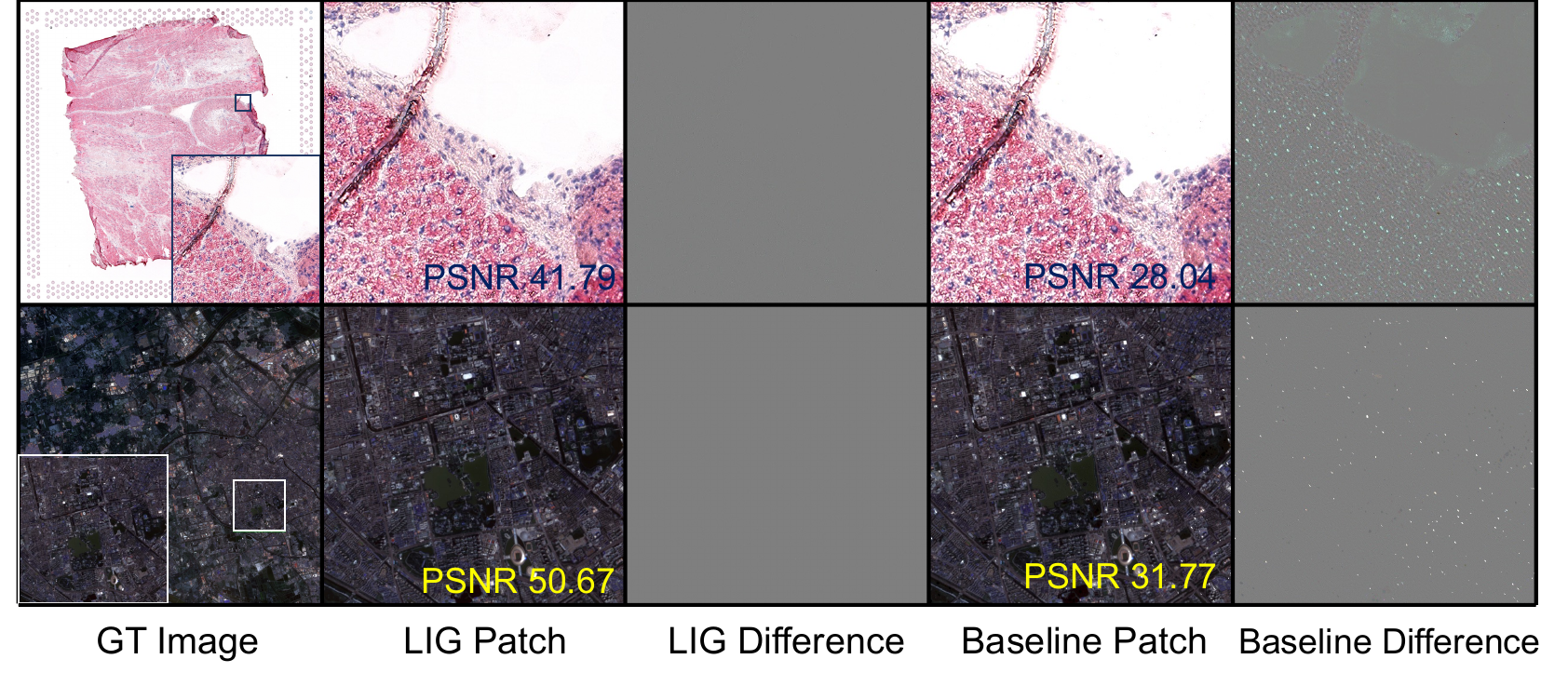}
\caption{\textbf{Qualitative comparison between LIG and GaussianImage on STimage and FGF2 samples.} We show small patches from the rendered images and the GT images. The difference images are shift to 0.5 for visualization.} 
\label{qualitative}
\end{figure*}

The quantitative results on three datasets are reported in Table~\ref{quantitative}.
We mainly compare our method with GaussianImage~\cite{zhang2024gaussianimage} focusing on large images, and also select INR-based methods for baselines. SIREN~\cite{sitzmann2020implicit}, Gauss~\cite{ramasinghe2022beyond}, WIRE~\cite{saragadam2023wire}, and Finers~\cite{liu2024finer} are selected for comparison. We report the quantitative results, including PSNR, Training Memory, and FPS, on three datasets. The FPS results are tested on the same environment. Note that these INRs are based on grid features, which suffer from large training memory requirements for large images since the batch sizes are too large. Therefore, we leave blank for those infeasible experiments. Additionally, while we can use tiny networks for running, the performances can be poor, as seen in the results of FINER where different network sizes are used for different datasets with training memory smaller than 80G. It is clear that compared with GS-based methods, INRs suffer from large training memory and low FPS.

Compared with GaussianImage, our method performs better on image quality, enabling GS for large signal fitting, especially for larger images on 4K and 9K. Regarding training memory, given the same number of Gaussian points, LIG does not propagate all the gradients but optimizes the levels in two stages, therefore reducing the training memory compared with our single level variant. From the Table ~\ref{quantitative}, we can also see that LIG consumes less memory than GaussianImage. Since the training and inference require two levels, the rendering speed can be slower than the GS-based baseline. However, as the additional level $L_0$ comprises fewer points and the final level $L_1$ has a reduced number of Gaussians, the FPS is not necessarily lower. For instance, on 9K images of STimage, with a total of 3.5e7 Gaussians, the FPS achieved is higher compared to GaussianImage. For other LIG results of the same point number, the reduction in FPS is mild considering the quality and training memory requirements. A qualitative comparison between LIG and GaussianImage is illustrated in Fig.~\ref{qualitative}. Note that for the histopathology image, the abundance of rich details may obscure the weaknesses of the baseline. Please refer to the difference images.

\subsection{Ablation Studies}
We present the ablation studies in Table~\ref{ablation}, evaluating the effectiveness of our two key components across different Gaussian point numbers on various datasets. All models are optimized for the same number of iterations. We utilize a variant representation of 2DGS and introduce a Level-of-Gaussian (LOG) mechanism. In this context, "w/o LOG" indicates the use of only the 2DGS variant with optimization performed at a single level, while "w/ LOG" refers to the full LIG implementation. The results clearly demonstrate that both components consistently yield performance improvements as the number of Gaussian points increases.

\begin{table}[t]
  \centering
  \resizebox{.48\textwidth}{!}{
  \begin{tabular}{lccccc}
    \toprule 
    Method & PSNR & & & & \\
    \midrule
    \textbf{STimage (9K)} & 2.5e7 & 3.5e7 & 4.5e7 & 5.5e7 & 6.5e7 \\
    \midrule
    GaussianImage & 31.86 & 29.86 & 29.33 & 29.28 & 29.26 \\
    Ours w/o LOG & 34.01 & 34.02 & 33.45 & 32.68 & 33.27 \\
    Ours w/ \enspace LOG & \textbf{35.03} & \textbf{37.47} & \textbf{39.82} & \textbf{42.19} & \textbf{44.49} \\
    \midrule
    \textbf{FGF2 (4K)} & 6e6 & 8e6 & 1e7 & 1.2e7 & 1.4e7 \\
    \midrule
    GaussianImage & 28.05 & 27.44 & 27.50 & 27.53 & 27.48 \\
    Ours w/o LOG & 47.07 & 48.54 & 49.65 & 50.50 & 51.35 \\
    Ours w/ \enspace LOG & \textbf{47.17} & \textbf{49.74} & \textbf{51.81} & \textbf{53.90} & \textbf{56.05} \\
    \midrule
    \textbf{DIV-HR (2K)} & 5e5 & 6e5 & 7e5 & 8e5 & 9e5 \\
    \midrule
    GaussianImage & 40.09 & 38.13 & 35.12 & 32.00 & 29.45 \\
    Ours w/o LOG & 43.21 & 43.66 & 43.47 & 42.97 & 42.30 \\
    Ours w/ \enspace LOG & \textbf{44.89} & \textbf{47.07} & \textbf{49.07} & \textbf{50.82} & \textbf{52.22} \\
    \bottomrule
  \end{tabular}
  }
  \caption{\textbf{Ablation studies on three datasets.} We evaluate the effectiveness of our two distinct designs compared to GaussianImage. Across various settings of Gaussian points, our designs consistently bring performance improvements.}
  \label{ablation}
\end{table}

\begin{table}[t]
  \centering
  \resizebox{0.48\textwidth}{!}{
  \begin{tabular}{lcccccc}
    \toprule 
   Setting & PSNR & FPS \\
    \midrule
    $|\mathcal{N}_0|, |\mathcal{N}_1| = 0, 30625000$ & 34.12 & \textbf{25.78} \\
    $|\mathcal{N}_0|, |\mathcal{N}_1| = 4375000, 30625000$ & \textbf{37.47} & 20.19 \\
    \midrule
    $|\mathcal{N}_0|, |\mathcal{N}_1| = 0, 39375000$ & 33.81 & \textbf{21.04} \\
    $|\mathcal{N}_0|, |\mathcal{N}_1| = 5625000, 39375000$ & \textbf{39.86} & 17.89 \\
    \midrule
    $|\mathcal{N}_0|, |\mathcal{N}_1| = 0, 48125000$ & 33.22 & \textbf{17.34} \\
    $|\mathcal{N}_0|, |\mathcal{N}_1| = 6875000, 48125000$ & \textbf{42.19} & 15.72 \\
    \bottomrule
  \end{tabular}
}
  \caption{\textbf{Effectiveness of low-frequency initialization for the second level on STimage.} We show comparison results with only using the number of Gaussian points in the second level for training a single level version of LIG.}
 \label{two-stage}
\end{table}

\begin{table}[h]
  \centering
  \resizebox{0.48\textwidth}{!}{
  \begin{tabular}{lcccccc}
    \toprule 
    Metrics / Iterations & 1e5 & 2e5 & 3e5 & 4e5 & 5e5 \\
    \midrule
    \textbf{STimage (9K)} & & & & &  \\
    \midrule
    PSNR (dB) $\uparrow$ & 37.68 & 39.12 & 39.82 & 40.38 & 40.76 \\
    Training Time (s) $\downarrow$ & 1825.68 & 3446.74 & 4777.65 & 6523.39 & 8028.46\\
    \midrule
    \textbf{FGF2 (4K)} & & & & &  \\
    \midrule
    PSNR (dB) $\uparrow$  & 53.22 & 55.11 & 56.05 & 56.70 & 57.20 \\
    Training Time (s) $\downarrow$ & 670.15 & 1306.56 & 1926.32 & 2562.91 & 3186.11 \\
    \bottomrule
  \end{tabular}
}
  \caption{\textbf{Performances with different training iterations.} The iteration number can be used to strike a balance between image quality and training time. The number for Gaussian points are (4.5e7, 1.4e7) for two datasets.}
  \label{iterations}
\end{table}

\begin{table}[h]
  \centering
  \resizebox{0.35\textwidth}{!}{
  \begin{tabular}{lccc}
    \toprule 
   Method & PSNR & Parameter Number & FPS \\
    \midrule
    LIG & 41.31 & 16000000 & 168.25\\
    NeuRBF & 44.69 & 17631486 & 156.38\\
    \bottomrule
  \end{tabular}
}
  \caption{\textbf{Comparison between LIG and NeuRBF on FGF2.} Given the similar total parameters for optimization, NeuRBF performs higher in PSNR while remains high FPS.}
 \label{neurbf}
\end{table}

\subsection{Further Analysis}
\textbf{Effectiveness of low-frequency initialization.} Our two-stage LOG approach benefits from easier training for large images due to the low-frequency initialization at the first level. In Table~\ref{two-stage}, we compare results with different point assignments to demonstrate the effectiveness of this initialization. We allocate additional points to the first level in the single-stage setting, with the only difference being the training target of the second level, denoted as $L_1$. This initialization results in higher PSNR, indicating that the training of the final target is facilitated. The two-stage process incurs extra inference time, leading to a slight drop in FPS.

\noindent \textbf{Trade-off between quality and training time.} One key factor affecting training time is the number of iterations. In Table~\ref{iterations}, we present results for different training iterations. The linearly increasing training time leads to higher fitting accuracy, with 3e5 iterations representing a compromise point adopted in our experiments.

\noindent \textbf{Comparison with NeuRBF.} We compare our method with the state-of-the-art INRs method, NeuRBF~\cite{chen2023neurbf}. In Table~\ref{neurbf}, we present performance metrics on the FGF2 dataset to illustrate the differences between our method and NeuRBF. We use 2e6 Gaussian points, comprising 1.6e7 parameters, to maintain a similar total number of parameters as NeuRBF for this dataset. It is observed that on 4K data, NeuRBF achieves higher PSNR while maintaining a similar FPS. With its radial bases and the solid foundation of Instant-NGP~\cite{muller2022instant}, NeuRBF advances INRs for large image fitting. While it remains unclear whether recent efficient techniques in INRs, such as hash coding, can be applied to 2DGS, existing and ongoing advancements in 3DGS may positively impact the development of novel 2DGS-based image representations. Given that GS-based image representation is still under-explored, we consider it to be in a very early stage.

\section{Conclusion}

In this work, we introduce Large Images are Gaussians (LIG) as a novel representation for large images. LIG is built upon 2D Gaussian Splatting (2DGS) and adopts a variant design for representing Gaussian points. Additionally, we propose a Level-of-Gaussian approach to facilitate the optimization of numerous 2D Gaussians. This enables 2DGS-based representation for fitting large images, significantly outperforming existing GS-based methods. While our work primarily focuses on representation and delves deeper into the performance of 2DGS as an image fitter, it is crucial to consider reducing the number of Gaussians for large images to achieve high-quality compression comparable to state-of-the-art Implicit Neural Representations (INRs).

\section{Acknowledgments}
This work was supported in part by the Research Grants Council of Hong Kong (27206123 and T45-401/22-N), in part by the Hong Kong Innovation and Technology Fund (ITS/273/22 and ITS/274/22), in part by the National Natural Science Foundation of China (No. 62201483), and in part by Guangdong Natural Science Fund (No. 2024A1515011875).

\bibliography{aaai25}

\begin{thebibliography}{55}
\providecommand{\natexlab}[1]{#1}

\bibitem[{Barron et~al.(2021)Barron, Mildenhall, Tancik, Hedman, Martin-Brualla, and Srinivasan}]{barron2021mip}
Barron, J.~T.; Mildenhall, B.; Tancik, M.; Hedman, P.; Martin-Brualla, R.; and Srinivasan, P.~P. 2021.
\newblock Mip-nerf: A multiscale representation for anti-aliasing neural radiance fields.
\newblock In \emph{Proceedings of the IEEE/CVF international conference on computer vision}, 5855--5864.

\bibitem[{Bengio, Courville, and Vincent(2013)}]{bengio2013representation}
Bengio, Y.; Courville, A.; and Vincent, P. 2013.
\newblock Representation learning: A review and new perspectives.
\newblock \emph{IEEE transactions on pattern analysis and machine intelligence}, 35(8): 1798--1828.

\bibitem[{Chen et~al.(2023{\natexlab{a}})Chen, Gwilliam, Lim, and Shrivastava}]{chen2023hnerv}
Chen, H.; Gwilliam, M.; Lim, S.-N.; and Shrivastava, A. 2023{\natexlab{a}}.
\newblock Hnerv: A hybrid neural representation for videos.
\newblock In \emph{Proceedings of the IEEE/CVF Conference on Computer Vision and Pattern Recognition}, 10270--10279.

\bibitem[{Chen et~al.(2021)Chen, He, Wang, Ren, Lim, and Shrivastava}]{chen2021nerv}
Chen, H.; He, B.; Wang, H.; Ren, Y.; Lim, S.~N.; and Shrivastava, A. 2021.
\newblock Nerv: Neural representations for videos.
\newblock \emph{Advances in Neural Information Processing Systems}, 34: 21557--21568.

\bibitem[{Chen et~al.(2024{\natexlab{a}})Chen, Li, Zhang, Zhu, Huang, Chen, and Lee}]{chen2024generalizable}
Chen, J.; Li, C.; Zhang, J.; Zhu, L.; Huang, B.; Chen, H.; and Lee, G.~H. 2024{\natexlab{a}}.
\newblock Generalizable Human Gaussians from Single-View Image.
\newblock \emph{arXiv preprint arXiv:2406.06050}.

\bibitem[{Chen et~al.(2024{\natexlab{b}})Chen, Zhou, Wu, Zhang, Li, and Li}]{chen2024stimage1k4m}
Chen, J.; Zhou, M.; Wu, W.; Zhang, J.; Li, Y.; and Li, D. 2024{\natexlab{b}}.
\newblock STimage-1K4M: A histopathology image-gene expression dataset for spatial transcriptomics.
\newblock arXiv:2406.06393.

\bibitem[{Chen, Liu, and Wang(2021)}]{chen2021learning}
Chen, Y.; Liu, S.; and Wang, X. 2021.
\newblock Learning continuous image representation with local implicit image function.
\newblock In \emph{Proceedings of the IEEE/CVF conference on computer vision and pattern recognition}, 8628--8638.

\bibitem[{Chen et~al.(2023{\natexlab{b}})Chen, Li, Song, Chen, Yu, Yuan, and Xu}]{chen2023neurbf}
Chen, Z.; Li, Z.; Song, L.; Chen, L.; Yu, J.; Yuan, J.; and Xu, Y. 2023{\natexlab{b}}.
\newblock Neurbf: A neural fields representation with adaptive radial basis functions.
\newblock In \emph{Proceedings of the IEEE/CVF International Conference on Computer Vision}, 4182--4194.

\bibitem[{Chen and Zhang(2019)}]{chen2019learning}
Chen, Z.; and Zhang, H. 2019.
\newblock Learning implicit fields for generative shape modeling.
\newblock In \emph{Proceedings of the IEEE/CVF conference on computer vision and pattern recognition}, 5939--5948.

\bibitem[{De~Sanctis et~al.(2015)De~Sanctis, Cianca, Araniti, Bisio, and Prasad}]{de2015satellite}
De~Sanctis, M.; Cianca, E.; Araniti, G.; Bisio, I.; and Prasad, R. 2015.
\newblock Satellite communications supporting internet of remote things.
\newblock \emph{IEEE Internet of Things Journal}, 3(1): 113--123.

\bibitem[{Dupont et~al.(2021)Dupont, Goli{\'n}ski, Alizadeh, Teh, and Doucet}]{dupont2021coin}
Dupont, E.; Goli{\'n}ski, A.; Alizadeh, M.; Teh, Y.~W.; and Doucet, A. 2021.
\newblock Coin: Compression with implicit neural representations.
\newblock \emph{arXiv preprint arXiv:2103.03123}.

\bibitem[{Hu et~al.(2024)Hu, Xia, Chen, Yang, and Zhang}]{hu2024gaussiansr}
Hu, J.; Xia, B.; Chen, B.; Yang, W.; and Zhang, L. 2024.
\newblock GaussianSR: High Fidelity 2D Gaussian Splatting for Arbitrary-Scale Image Super-Resolution.
\newblock \emph{arXiv preprint arXiv:2407.18046}.

\bibitem[{Huang et~al.(2024)Huang, Yu, Chen, Geiger, and Gao}]{huang20242d}
Huang, B.; Yu, Z.; Chen, A.; Geiger, A.; and Gao, S. 2024.
\newblock 2d gaussian splatting for geometrically accurate radiance fields.
\newblock In \emph{ACM SIGGRAPH 2024 Conference Papers}, 1--11.

\bibitem[{Kerbl et~al.(2023)Kerbl, Kopanas, Leimk{\"u}hler, and Drettakis}]{kerbl20233d}
Kerbl, B.; Kopanas, G.; Leimk{\"u}hler, T.; and Drettakis, G. 2023.
\newblock 3D Gaussian Splatting for Real-Time Radiance Field Rendering.
\newblock \emph{ACM Trans. Graph.}, 42(4): 139--1.

\bibitem[{Kerbl et~al.(2024)Kerbl, Meuleman, Kopanas, Wimmer, Lanvin, and Drettakis}]{kerbl2024hierarchical}
Kerbl, B.; Meuleman, A.; Kopanas, G.; Wimmer, M.; Lanvin, A.; and Drettakis, G. 2024.
\newblock A hierarchical 3d gaussian representation for real-time rendering of very large datasets.
\newblock \emph{ACM Transactions on Graphics (TOG)}, 43(4): 1--15.

\bibitem[{Khayam(2003)}]{khayam2003discrete}
Khayam, S.~A. 2003.
\newblock The discrete cosine transform (DCT): theory and application.
\newblock \emph{Michigan State University}, 114(1): 31.

\bibitem[{Kingma and Ba(2014)}]{kingma2014adam}
Kingma, D.~P.; and Ba, J. 2014.
\newblock Adam: A method for stochastic optimization.
\newblock \emph{arXiv preprint arXiv:1412.6980}.

\bibitem[{Lassner and Zollhofer(2021)}]{lassner2021pulsar}
Lassner, C.; and Zollhofer, M. 2021.
\newblock Pulsar: Efficient sphere-based neural rendering.
\newblock In \emph{Proceedings of the IEEE/CVF Conference on Computer Vision and Pattern Recognition}, 1440--1449.

\bibitem[{LeCun, Bengio, and Hinton(2015)}]{lecun2015deep}
LeCun, Y.; Bengio, Y.; and Hinton, G. 2015.
\newblock Deep learning.
\newblock \emph{nature}, 521(7553): 436--444.

\bibitem[{Li et~al.(2024)Li, Feng, Liu, Liu, Wang, Yu, and Yuan}]{li2024endosparse}
Li, C.; Feng, B.~Y.; Liu, Y.; Liu, H.; Wang, C.; Yu, W.; and Yuan, Y. 2024.
\newblock Endosparse: Real-time sparse view synthesis of endoscopic scenes using gaussian splatting.
\newblock In \emph{International Conference on Medical Image Computing and Computer-Assisted Intervention}, 252--262. Springer.

\bibitem[{Li et~al.(2022)Li, Wang, Pi, Xu, Mei, and Liu}]{li2022nerv}
Li, Z.; Wang, M.; Pi, H.; Xu, K.; Mei, J.; and Liu, Y. 2022.
\newblock E-nerv: Expedite neural video representation with disentangled spatial-temporal context.
\newblock In \emph{European Conference on Computer Vision}, 267--284. Springer.

\bibitem[{Liu et~al.(2024{\natexlab{a}})Liu, Zhan, Tang, Shan, Zeng, Lin, Liu, and Liu}]{liu2024humangaussian}
Liu, X.; Zhan, X.; Tang, J.; Shan, Y.; Zeng, G.; Lin, D.; Liu, X.; and Liu, Z. 2024{\natexlab{a}}.
\newblock Humangaussian: Text-driven 3d human generation with gaussian splatting.
\newblock In \emph{Proceedings of the IEEE/CVF Conference on Computer Vision and Pattern Recognition}, 6646--6657.

\bibitem[{Liu et~al.(2024{\natexlab{b}})Liu, Li, Yang, and Yuan}]{liu2024endogaussian}
Liu, Y.; Li, C.; Yang, C.; and Yuan, Y. 2024{\natexlab{b}}.
\newblock Endogaussian: Gaussian splatting for deformable surgical scene reconstruction.
\newblock \emph{arXiv preprint arXiv:2401.12561}.

\bibitem[{Liu et~al.(2024{\natexlab{c}})Liu, Zhu, Zhang, Fu, Deng, Ma, Guo, and Cao}]{liu2024finer}
Liu, Z.; Zhu, H.; Zhang, Q.; Fu, J.; Deng, W.; Ma, Z.; Guo, Y.; and Cao, X. 2024{\natexlab{c}}.
\newblock FINER: Flexible spectral-bias tuning in Implicit NEural Representation by Variable-periodic Activation Functions.
\newblock In \emph{Proceedings of the IEEE/CVF Conference on Computer Vision and Pattern Recognition}, 2713--2722.

\bibitem[{Lu et~al.(2024)Lu, Zhang, Wang, Liu, Lu, and Tang}]{lu2024manigaussian}
Lu, G.; Zhang, S.; Wang, Z.; Liu, C.; Lu, J.; and Tang, Y. 2024.
\newblock Manigaussian: Dynamic gaussian splatting for multi-task robotic manipulation.
\newblock \emph{arXiv preprint arXiv:2403.08321}.

\bibitem[{Ma et~al.(2022)Ma, Yu, Lu, and Zhou}]{ma2022recovering}
Ma, C.; Yu, P.; Lu, J.; and Zhou, J. 2022.
\newblock Recovering realistic details for magnification-arbitrary image super-resolution.
\newblock \emph{IEEE Transactions on Image Processing}, 31: 3669--3683.

\bibitem[{Martel et~al.(2021)Martel, Lindell, Lin, Chan, Monteiro, and Wetzstein}]{martel2021acorn}
Martel, J.~N.; Lindell, D.~B.; Lin, C.~Z.; Chan, E.~R.; Monteiro, M.; and Wetzstein, G. 2021.
\newblock Acorn: Adaptive coordinate networks for neural scene representation.
\newblock \emph{arXiv preprint arXiv:2105.02788}.

\bibitem[{Matsuki et~al.(2024)Matsuki, Murai, Kelly, and Davison}]{matsuki2024gaussian}
Matsuki, H.; Murai, R.; Kelly, P.~H.; and Davison, A.~J. 2024.
\newblock Gaussian splatting slam.
\newblock In \emph{Proceedings of the IEEE/CVF Conference on Computer Vision and Pattern Recognition}, 18039--18048.

\bibitem[{Michalkiewicz et~al.(2019)Michalkiewicz, Pontes, Jack, Baktashmotlagh, and Eriksson}]{michalkiewicz2019implicit}
Michalkiewicz, M.; Pontes, J.~K.; Jack, D.; Baktashmotlagh, M.; and Eriksson, A. 2019.
\newblock Implicit surface representations as layers in neural networks.
\newblock In \emph{Proceedings of the IEEE/CVF International Conference on Computer Vision}, 4743--4752.

\bibitem[{Mildenhall et~al.(2021)Mildenhall, Srinivasan, Tancik, Barron, Ramamoorthi, and Ng}]{mildenhall2021nerf}
Mildenhall, B.; Srinivasan, P.~P.; Tancik, M.; Barron, J.~T.; Ramamoorthi, R.; and Ng, R. 2021.
\newblock Nerf: Representing scenes as neural radiance fields for view synthesis.
\newblock \emph{Communications of the ACM}, 65(1): 99--106.

\bibitem[{Mittermaier, Venkatesh, and Kvedar(2023)}]{mittermaier2023digital}
Mittermaier, M.; Venkatesh, K.~P.; and Kvedar, J.~C. 2023.
\newblock Digital health technology in clinical trials.
\newblock \emph{NPJ Digital Medicine}, 6(1): 88.

\bibitem[{M{\"u}ller et~al.(2022)M{\"u}ller, Evans, Schied, and Keller}]{muller2022instant}
M{\"u}ller, T.; Evans, A.; Schied, C.; and Keller, A. 2022.
\newblock Instant neural graphics primitives with a multiresolution hash encoding.
\newblock \emph{ACM transactions on graphics (TOG)}, 41(4): 1--15.

\bibitem[{Park et~al.(2019)Park, Florence, Straub, Newcombe, and Lovegrove}]{park2019deepsdf}
Park, J.~J.; Florence, P.; Straub, J.; Newcombe, R.; and Lovegrove, S. 2019.
\newblock Deepsdf: Learning continuous signed distance functions for shape representation.
\newblock In \emph{Proceedings of the IEEE/CVF conference on computer vision and pattern recognition}, 165--174.

\bibitem[{Rabbani and Joshi(2002)}]{rabbani2002overview}
Rabbani, M.; and Joshi, R. 2002.
\newblock An overview of the JPEG 2000 still image compression standard.
\newblock \emph{Signal processing: Image communication}, 17(1): 3--48.

\bibitem[{Ramasinghe and Lucey(2022)}]{ramasinghe2022beyond}
Ramasinghe, S.; and Lucey, S. 2022.
\newblock Beyond periodicity: Towards a unifying framework for activations in coordinate-mlps.
\newblock In \emph{European Conference on Computer Vision}, 142--158. Springer.

\bibitem[{Ren et~al.(2024)Ren, Jiang, Lu, Yu, Xu, Ni, and Dai}]{ren2024octree}
Ren, K.; Jiang, L.; Lu, T.; Yu, M.; Xu, L.; Ni, Z.; and Dai, B. 2024.
\newblock Octree-gs: Towards consistent real-time rendering with lod-structured 3d gaussians.
\newblock \emph{arXiv preprint arXiv:2403.17898}.

\bibitem[{Saragadam et~al.(2023)Saragadam, LeJeune, Tan, Balakrishnan, Veeraraghavan, and Baraniuk}]{saragadam2023wire}
Saragadam, V.; LeJeune, D.; Tan, J.; Balakrishnan, G.; Veeraraghavan, A.; and Baraniuk, R.~G. 2023.
\newblock Wire: Wavelet implicit neural representations.
\newblock In \emph{Proceedings of the IEEE/CVF Conference on Computer Vision and Pattern Recognition}, 18507--18516.

\bibitem[{Saragadam et~al.(2022)Saragadam, Tan, Balakrishnan, Baraniuk, and Veeraraghavan}]{saragadam2022miner}
Saragadam, V.; Tan, J.; Balakrishnan, G.; Baraniuk, R.~G.; and Veeraraghavan, A. 2022.
\newblock Miner: Multiscale implicit neural representation.
\newblock In \emph{European Conference on Computer Vision}, 318--333. Springer.

\bibitem[{Sitzmann et~al.(2020)Sitzmann, Martel, Bergman, Lindell, and Wetzstein}]{sitzmann2020implicit}
Sitzmann, V.; Martel, J.; Bergman, A.; Lindell, D.; and Wetzstein, G. 2020.
\newblock Implicit neural representations with periodic activation functions.
\newblock \emph{Advances in neural information processing systems}, 33: 7462--7473.

\bibitem[{Str{\"u}mpler et~al.(2022)Str{\"u}mpler, Postels, Yang, Gool, and Tombari}]{strumpler2022implicit}
Str{\"u}mpler, Y.; Postels, J.; Yang, R.; Gool, L.~V.; and Tombari, F. 2022.
\newblock Implicit neural representations for image compression.
\newblock In \emph{European Conference on Computer Vision}, 74--91. Springer.

\bibitem[{Szymanowicz, Rupprecht, and Vedaldi(2024)}]{szymanowicz2024splatter}
Szymanowicz, S.; Rupprecht, C.; and Vedaldi, A. 2024.
\newblock Splatter image: Ultra-fast single-view 3d reconstruction.
\newblock In \emph{Proceedings of the IEEE/CVF Conference on Computer Vision and Pattern Recognition}, 10208--10217.

\bibitem[{Tang et~al.(2024)Tang, Chen, Chen, Wang, Zeng, and Liu}]{tang2024lgm}
Tang, J.; Chen, Z.; Chen, X.; Wang, T.; Zeng, G.; and Liu, Z. 2024.
\newblock Lgm: Large multi-view gaussian model for high-resolution 3d content creation.
\newblock \emph{arXiv preprint arXiv:2402.05054}.

\bibitem[{Tang et~al.(2023)Tang, Ren, Zhou, Liu, and Zeng}]{tang2023dreamgaussian}
Tang, J.; Ren, J.; Zhou, H.; Liu, Z.; and Zeng, G. 2023.
\newblock Dreamgaussian: Generative gaussian splatting for efficient 3d content creation.
\newblock \emph{arXiv preprint arXiv:2309.16653}.

\bibitem[{Wang et~al.(2024)Wang, He, Dong, Lin, Huang, and Ding}]{wang2024cross}
Wang, Y.; He, X.; Dong, Y.; Lin, Y.; Huang, Y.; and Ding, X. 2024.
\newblock Cross-Modality Interaction Network for Pan-sharpening.
\newblock \emph{IEEE Transactions on Geoscience and Remote Sensing}.

\bibitem[{Xiangli et~al.(2022)Xiangli, Xu, Pan, Zhao, Rao, Theobalt, Dai, and Lin}]{xiangli2022bungeenerf}
Xiangli, Y.; Xu, L.; Pan, X.; Zhao, N.; Rao, A.; Theobalt, C.; Dai, B.; and Lin, D. 2022.
\newblock Bungeenerf: Progressive neural radiance field for extreme multi-scale scene rendering.
\newblock In \emph{European conference on computer vision}, 106--122. Springer.

\bibitem[{Xu et~al.(2019)Xu, Wang, Ceylan, Mech, and Neumann}]{xu2019disn}
Xu, Q.; Wang, W.; Ceylan, D.; Mech, R.; and Neumann, U. 2019.
\newblock Disn: Deep implicit surface network for high-quality single-view 3d reconstruction.
\newblock \emph{Advances in neural information processing systems}, 32.

\bibitem[{Xu et~al.(2024)Xu, Shi, Yifan, Chen, Yang, Peng, Shen, and Wetzstein}]{xu2024grm}
Xu, Y.; Shi, Z.; Yifan, W.; Chen, H.; Yang, C.; Peng, S.; Shen, Y.; and Wetzstein, G. 2024.
\newblock Grm: Large gaussian reconstruction model for efficient 3d reconstruction and generation.
\newblock \emph{arXiv preprint arXiv:2403.14621}.

\bibitem[{Yan et~al.(2024)Yan, Low, Chen, and Lee}]{yan2024multi}
Yan, Z.; Low, W.~F.; Chen, Y.; and Lee, G.~H. 2024.
\newblock Multi-scale 3d gaussian splatting for anti-aliased rendering.
\newblock In \emph{Proceedings of the IEEE/CVF Conference on Computer Vision and Pattern Recognition}, 20923--20931.

\bibitem[{Ye and Kanazawa(2023)}]{ye2023mathematical}
Ye, V.; and Kanazawa, A. 2023.
\newblock Mathematical Supplement for the $\texttt{gsplat}$ Library.
\newblock arXiv:2312.02121.

\bibitem[{Yu et~al.(2024)Yu, Chen, Huang, Sattler, and Geiger}]{yu2024mip}
Yu, Z.; Chen, A.; Huang, B.; Sattler, T.; and Geiger, A. 2024.
\newblock Mip-splatting: Alias-free 3d gaussian splatting.
\newblock In \emph{Proceedings of the IEEE/CVF Conference on Computer Vision and Pattern Recognition}, 19447--19456.

\bibitem[{Yugay et~al.(2023)Yugay, Li, Gevers, and Oswald}]{yugay2023gaussian}
Yugay, V.; Li, Y.; Gevers, T.; and Oswald, M.~R. 2023.
\newblock Gaussian-slam: Photo-realistic dense slam with gaussian splatting.
\newblock \emph{arXiv preprint arXiv:2312.10070}.

\bibitem[{Zhang et~al.(2024{\natexlab{a}})Zhang, Ge, Xu, He, Wang, Qin, Lu, Geng, and Zhang}]{zhang2024gaussianimage}
Zhang, X.; Ge, X.; Xu, T.; He, D.; Wang, Y.; Qin, H.; Lu, G.; Geng, J.; and Zhang, J. 2024{\natexlab{a}}.
\newblock GaussianImage: 1000 FPS Image Representation and Compression by 2D Gaussian Splatting.
\newblock \emph{arXiv preprint arXiv:2403.08551}.

\bibitem[{Zhang et~al.(2024{\natexlab{b}})Zhang, Kuznetsov, Jindal, Chen, Sochenov, Kaplanyan, and Sun}]{zhang2024image}
Zhang, Y.; Kuznetsov, A.; Jindal, A.; Chen, K.; Sochenov, A.; Kaplanyan, A.; and Sun, Q. 2024{\natexlab{b}}.
\newblock Image-GS: Content-Adaptive Image Representation via 2D Gaussians.
\newblock \emph{arXiv preprint arXiv:2407.01866}.

\bibitem[{Zhao et~al.(2024)Zhao, Zhao, Zhu, Zheng, and Xu}]{zhao2024hfgs}
Zhao, H.; Zhao, X.; Zhu, L.; Zheng, W.; and Xu, Y. 2024.
\newblock HFGS: 4D Gaussian Splatting with Emphasis on Spatial and Temporal High-Frequency Components for Endoscopic Scene Reconstruction.
\newblock \emph{arXiv preprint arXiv:2405.17872}.

\bibitem[{Zhu et~al.(2024)Zhu, Wang, Jin, Lin, and Yu}]{zhu2024deformable}
Zhu, L.; Wang, Z.; Jin, Z.; Lin, G.; and Yu, L. 2024.
\newblock Deformable endoscopic tissues reconstruction with gaussian splatting.
\newblock \emph{arXiv preprint arXiv:2401.11535}.

\end{thebibliography}

\end{document}